\documentclass[letterpaper]{article} 
\usepackage{aaai24}  
\usepackage{times}  
\usepackage{helvet}  
\usepackage{courier}  
\usepackage[hyphens]{url}  
\usepackage{graphicx} 
\usepackage{natbib}  
\usepackage{caption} 
\usepackage{algorithm}
\usepackage{algorithmic}

\usepackage{multirow}
\usepackage{amsthm,amsmath,amssymb}
\usepackage{mathrsfs}
\usepackage{graphicx}
\usepackage{amsmath}
\usepackage{amssymb}
\usepackage{booktabs}
\usepackage{times}
\usepackage{microtype}
\usepackage{epsfig}
\usepackage{placeins}

\newcommand{\et}[2]{${#1}^{\pm{#2}}$}
\newcommand{\etb}[2]{$\mathbf{{#1}}^{\pm{#2}}$}
\newcommand{\etr}[2]{$\textcolor{red}{{#1}}^{\pm{#2}}$}
\newcommand{\etbb}[2]{$\textcolor{blue}{{#1}}^{\pm{#2}}$}

\usepackage{xcolor}

\usepackage{newfloat}
\usepackage{listings}
\DeclareCaptionStyle{ruled}{labelfont=normalfont,labelsep=colon,strut=off} 
\floatstyle{ruled}
\newfloat{listing}{tb}{lst}{}
\floatname{listing}{Listing}
\title{Towards Detailed Text-to-Motion Synthesis via Basic-to-Advanced \\ Hierarchical Diffusion Model}
\author{
    Zhenyu Xie\textsuperscript{\rm 1}, Yang Wu\textsuperscript{\rm 2}, Xuehao Gao\textsuperscript{\rm 3}, \\
    Zhongqian Sun\textsuperscript{\rm 2}, Wei Yang\textsuperscript{\rm 2}, Xiaodan Liang\textsuperscript{\rm 1}\thanks{Xiaodan Liang is the corresponding author. Code is available at https://github.com/xiezhy6/B2A-HDM.}
}
\affiliations{
    \textsuperscript{\rm 1}Shenzhen Campus of Sun Yat-sen University, \textsuperscript{\rm 2}Tencent AI Lab,
    \textsuperscript{\rm 3} Xi'an Jiao Tong University \\

    xiezhy6@mail2.sysu.edu.cn, dylan.yangwu@qq.com, gaoxuehao.xjtu@gmail.com, \\
    sallensun@tencent.com, willyang@tencent.com, xdliang328@gmail.com
}

\usepackage{bibentry}

\begin{document}

\maketitle

\begin{abstract}
Text-guided motion synthesis aims to generate 3D human motion that not only precisely reflects the textual description but reveals the motion details as much as possible. Pioneering methods explore the diffusion model for text-to-motion synthesis and obtain significant superiority. However, these methods conduct diffusion processes either on the raw data distribution or the low-dimensional latent space, which typically suffer from the problem of modality inconsistency or detail-scarce. To tackle this problem, we propose a novel \textbf{B}asic-to-\textbf{A}dvanced \textbf{H}ierarchical \textbf{D}iffusion \textbf{M}odel, named B2A-HDM, to collaboratively exploit low-dimensional and  high-dimensional diffusion models for high quality detailed motion synthesis. Specifically, the basic diffusion model in low-dimensional latent space provides the intermediate denoising result that to be consistent with the textual description, while the advanced diffusion model in high-dimensional latent space focuses on the following detail-enhancing denoising process. 
Besides, we introduce a multi-denoiser framework for the advanced diffusion model to ease the learning of high-dimensional model and fully explore the generative potential of the diffusion model. Quantitative and qualitative experiment results on two text-to-motion benchmarks (HumanML3D and KIT-ML) demonstrate that B2A-HDM can outperform existing state-of-the-art methods in terms of fidelity, modality consistency, and diversity.
\end{abstract}

\begin{figure*}[t]
  \centering
  \includegraphics[width=1.0\hsize]{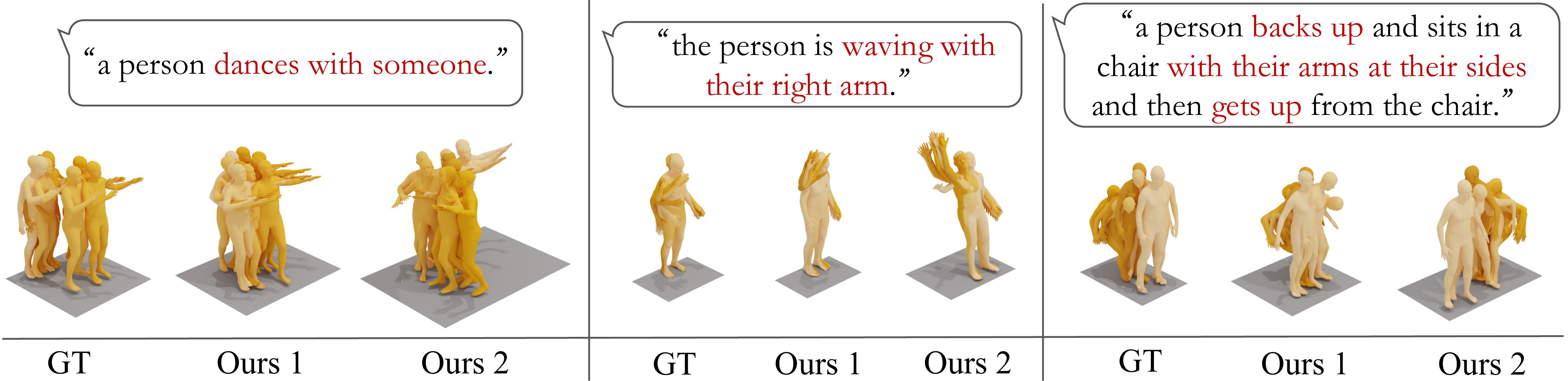}
  \vspace{-4mm}
  \caption{The visual results of our B2A-HDM on HumanML3D~\cite{guo2022t2m}. Our method can generate diverse and high-quality motion sequences that conform to the provided textual descriptions.}
   \vspace{-2mm}
  \label{fig:teaser}
\end{figure*}

\begin{figure*}[t]
  \centering
  \includegraphics[width=1.0\hsize]{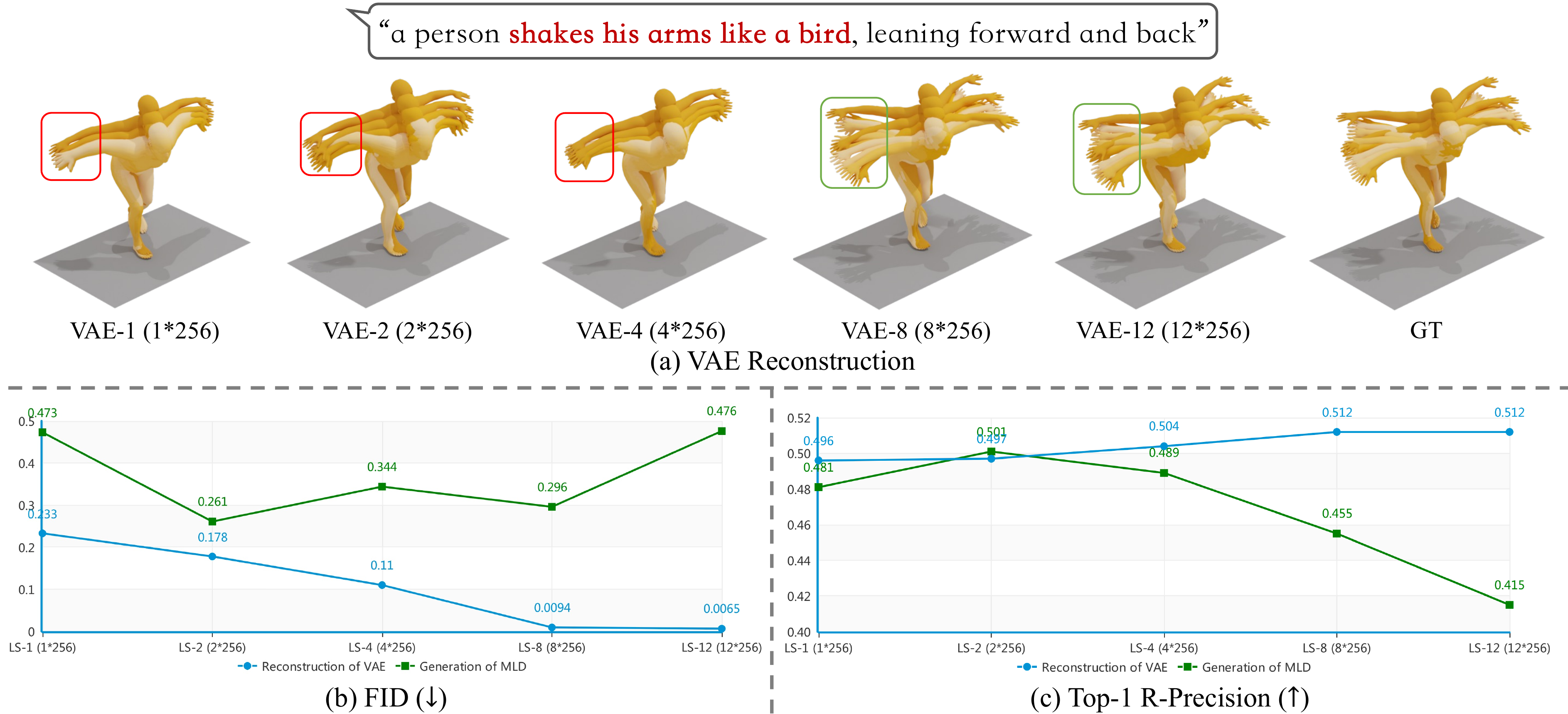}
  \vspace{-6mm}
  \caption{(a) Visual comparisons among the reconstruction results of different VAEs. (b) Comparison of FID scores (lower is better). (c) Comparison of Top-1 R-Precision scores (higher is better).}
   \vspace{-6mm}
  \label{fig:dimension_comps}
\end{figure*}

\section{Introduction}
Text-to-motion synthesis, which aims to generate human motion that conforms to the textual descriptions (with result examples of our proposed model shown in Figure \ref{fig:teaser}), has made significant progress in recent years. It has the potential to revolutionize the traditional process of acquiring human motion, which typically requires expert knowledge from artists or expensive motion capture equipment. 

However, inferring human motion from its textual description is a no-trivial task due to the essential discrepancy between the two data modalities. To address this challenge, some existing works~\cite{ahuja2019l2p,ghosh2021hier,tevet2022motionclip,petrovich22temos} resort to the auto-encoder/VAE for motion synthesis and strive to align the cross-modal information in a shared embedding space.
On the other hand, motivated by fruitful attempts of diffusion model~\cite{dickstein2015dm,ho2020denoising,nichol2021} in the cross-modal image synthesis~\cite{rombach2021highresolution,saharia2022imagen,ramesh2022dalle2}, some pioneering works~\cite{tevet2023mdm,zhang2022motiondiffuse,dabral2022mofusion,ma2022mofusion,chen2023mld,jin2023act} exploit the diffusion model for text-to-motion synthesis, demonstrating significant improvements in fidelity and cross-modal consistency.

In spite of the powerful generative ability, training a diffusion model for text-to-motion synthesis remains challenging, which is mainly attributed to the complexity of the data distribution and the insufficiency of the text-annotated training data. Without adequate data, it is difficult for the neural network to learn the denoising process that converts the Gaussian distribution into the complex motion distribution.
To address this problem, MLD~\cite{chen2023mld} applies VAE to project the raw motion from the initial 3D pose space into the latent code in the low-dimensional latent space, and then conducts diffusion process on the latent space.
Although simplifying the target distribution can ease the learning of the denoising process, the low-dimensional latent code is less expressive, which hinders the diffusion model from generating detailed motion.
Specifically, as shown in Fig.~\ref{fig:dimension_comps}(a), reducing the dimension of the latent space in VAE results in reconstructed motions with fewer captured details. Additionally, Fig.~\ref{fig:dimension_comps}(b) reveals that decreasing the dimension of the latent space leads to an increase in the FID scores of the reconstructed motions, indicating a degradation in their quality.
However, simply increasing the dimension of the latent space makes the target distribution complex again, leading to more difficulties in network learning, which will further result in a significant drop in performance in terms of cross-modal consistency, as demonstrated in Fig.~\ref{fig:dimension_comps}(c).
Nevertheless, the comparisons in Fig.~\ref{fig:dimension_comps}(c) provide an intuitive insight that while the low-dimensional diffusion model is ineffective for detail generation, it significantly benefits the modality transformation.
This insight further inspires us to integrate the complementary advantages of low-dimensional and high-dimensional diffusion models to enhance cross-modal consistency and facilitate detail-rich motion generation.

To this end, we proposed a novel \textbf{B}asic-to-\textbf{A}dvanced \textbf{H}ierarchical \textbf{D}iffusion \textbf{M}odel, named B2A-HDM, for text-guided motion synthesis, in which the basic diffusion model focuses on consistent but detail-scarce text-to-motion transformation, 
while the advanced diffusion model aims to conduct a detail-enhancing denoising process based on the intermediate results from the basic model. 

Specifically, the Basic Diffusion Model (BDM) is trained in the low-dimensional latent space, in which the data distribution is much simpler than the raw motion distribution, making the learning of text-to-motion transformation easier.
However, since the low-dimensional latent space is less expressive, BDM is ineffective to synthesize detail-rich results.
On the other hand, the Advanced Diffusion Model (ADM) is trained on the high-dimensional latent space, thus has a larger representation capacity for characterizing more motion details and improving high-fidelity synthesis. 
However, directly using ADM to conduct the whole text-to-motion denoising process will lead to poor modality consistency. To tackle this problem, B2A-HDM explicitly divides the denoising process into several sub-processes, in which BDM and ADM focus on different denoising stages.
To be specific, B2A-HDM first conducts the forward diffusion on the synthesized result from BDM, resulting in the noised motion that will be regarded as the result of the early denoising sub-process. Then, ADM conducts the following denoising process based on the noised motion.
Since the noised motion derived from BDM provides a proper initial state (i.e., consistent with the textual description), ADM can focus on the detail-enhancing denoising process.
Moreover, to further ease the learning of high-dimensional diffusion model, B2A-HDM exploits the multi-denoisers framework for ADM, in which each denoiser dominates a specific denoising sub-process.

Overall, our contributions can be summarized as follows: (1) We propose a novel Basic-to-Advanced Hierarchical Diffusion Model (B2A-HDM) for text-to-motion synthesis, which jointly incorporates the complementary benefits of low-/high-dimensional diffusion models into detailed motion synthesis. 
(2) We explicitly divides the denoising process into several sub-processes, which are separately dominated by one basic and two advanced diffusion models.
(3) Extensive experiments on two text-to-motion benchmarks~\cite{guo2022t2m,Plappert2016} show the superiority of B2A-HDM over the existing SOTAs.

\section{Related Work}
\noindent\textbf{Conditional Diffusion Models.}
Diffusion models~\cite{dickstein2015dm,ho2020denoising,nichol2021} are a novel class of generative models that have made significant strides in cross-modal synthesis, spanning a diverse range of applications such as text-to-image~\cite{rombach2021highresolution,saharia2022imagen,ramesh2022dalle2}, text-to-video~\cite{ho2022imagenvideo,esser2023gen1,yu2023video}, text-to-3d~\cite{poole2022dreamfusion,lin2023magic3d}, text-to-audio~\cite{popv2021tts}, among others.
Typically, diffusion models consist of the forward and reverse diffusion process, in which the forward process gradually add Gaussian noise into real data to construct training data, while the reverse process involves a neural network to conduct denoising.
To adapt diffusion models for conditional generation, Dharival et al.~\cite{dharival2021dbg} propose a classifier-guided diffusion model that incorporates conditional information into the reverse diffusion process using additional classifiers. Besides, Ho et al.~\cite{ho2021cfg} propose a classifier-free guidance strategy for conditional diffusion models. 
This approach strikes a balance between synthesis quality and diversity, which can obtain better results and is widely used by the following works.

\noindent\textbf{Text-Guided Human Motion Synthesis.}
Following the development of generative models, text-guided motion synthesis has witnessed significant progress in recent years. JL2P~\cite{ahuja2019l2p} employs auto-encoder to model a share space for the text and motion embedding, from which the text embedding will be used to reconstruct the corresponding motion during inference.
MotionCLIP~\cite{tevet2022motionclip} attempts to improve the auto-encoder's generalization by aligning the shared space with the expressive CLIP~\cite{nichol2021clip} embedding space, enabling it to handle out-of-distribution motion synthesis.
TMEOS~\cite{petrovich22temos} and T2M~\cite{guo2022t2m} use VAE framework to enhance the diversity of the generated results by constraining the share space into a normal distribution. 
Different from the above encoder-decoder paradigm, T2M-GPT~\cite{zhang2023generating} generates motion sequence in an auto-regressive manner by jointly using VQ-VAE~\cite{oord2017vqvae} and GPT~\cite{radford2018gpt}, which gains improvement in term of fidelity and modality consistency. 
Building on the great success of diffusion model on image synthesis, some recent works~\cite{tevet2023mdm,zhang2022motiondiffuse,ma2022mofusion,chen2023mld} explore the generative potential of diffusion for text-to-motion synthesis.
However, these methods model the diffusion process either on the raw motion distribution or on a low-dimensional latent space, leading to modality-inconsistent or detail-scarce synthesis. In this paper, we exploit the basic and advanced diffusion models in different latent spaces to conduct the reverse diffusion process, in which basic and advanced models are separately in charge of modality transformation and detail-enhancing denoising process. Note that, while eDiff-I~\cite{balaji2022eDiff-I}  also employs multiple denoisers for reverse diffusion, our method differs in that we we train denoisers on different latent spaces, whereas in eDiff-I  various denoisers are all modeled on the same raw data space.

\section{Methodology}

\begin{figure*}[t]
  \centering
  \includegraphics[width=1.0\hsize]{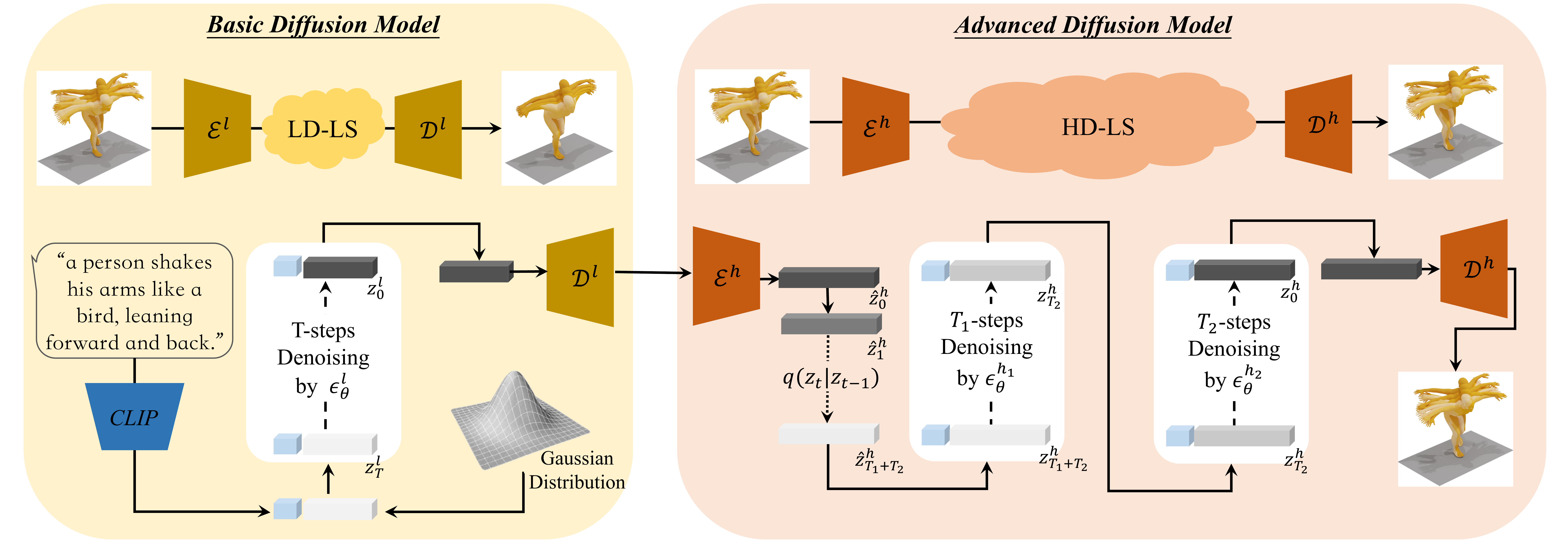}
  \vspace{-4mm}
  \caption{Method Overview. B2A-HDM consists of a Basic Diffusion Model(BDM) and an Advanced Diffusion Model(ADM). BDM comprises a VAE $\{ \mathcal{E}_l, \mathcal{D}_l \}$ and a denoiser $\epsilon_{\theta}^l$, in which $\epsilon_{\theta}^l$ is in charge of the complete $T$-steps denoising process in the low-dimensional latent space (LD-LS). ADM comprises a VAE $\{ \mathcal{E}_h, \mathcal{D}_h \}$ and two denoisers $\epsilon_{\theta}^{h_1}$ and $\epsilon_{\theta}^{h_2}$, in which $\epsilon_{\theta}^{h_1}$ and $\epsilon_{\theta}^{h_2}$ are responsible for $T_1$- and $T_2$-steps denoising sub-process on the high-dimensional latent space (HD-LS), respectively.}
   \vspace{-5mm}
  \label{fig:framework}
\end{figure*}

Given a textual description $\mathbf{w}=\left\{w_i\right\}_{i=1}^L$ with $L$ words,  text-to-motion synthesis aims to generate the 3D motion $\mathbf{s}=\left\{s_i\right\}_{i=1}^N$ 
with $N$ frames that conform to the text input, where $s_i \in \mathbb{R}^{J}$ denotes a $J$-dimensional body pose representation at $i$-th frame.
To achieve this, we propose a novel Basic-to-Advanced Hierarchical Diffusion Model (B2A-HDM) to collaboratively exploit the low- and high-dimensional latent diffusion models for modality consistency and detail-rich motion synthesis.
In the following, we will first introduce the latent diffusion model for text-to-motion synthesis in Sec.~\ref{sec3.1:ldm}. Then, we will explain the technical details of our B2A-HDM in Sec.~\ref{sec3.2:b2ahdm}. Finally, we will discuss the training details and objective functions in Sec.~\ref{sec3.3:train-test}. A method overview is shown in Fig.~\ref{fig:framework}.

\subsection{Latent Diffusion for Text-to-Motion Synthesis}\label{sec3.1:ldm}
Recently, diffusion model~\cite{dickstein2015dm,ho2020denoising,nichol2021} has shown its outstanding generative ability for cross-modal synthesis tasks~\cite{rombach2021highresolution,saharia2022imagen,ramesh2022dalle2,esser2023gen1}, inspiring researchers to explore diffusion model for high-quality text-to-motion synthesis. 
However, directly modeling the diffusion process on the raw motion distribution typically suffers from inferior synthesis due to the high complexity of raw distribution and the insufficiency of text-annotated training data. To tackle this problem, existing method~\cite{chen2023mld} exploits a latent diffusion model to degrade the complexity of target distribution and conduct the diffusion process in low-dimensional latent space, leading to higher quality text-to-motion synthesis.

Specifically, the latent diffusion model is composed of a motion VAE and a diffusion model in VAE latent space. 
The motion VAE consists of transformer-based encoder $\mathcal{E}$ and decoder $\mathcal{D}$, in which $\mathcal{E}$ is used to encode the raw motion $\mathbf{s} \in \mathbb{R}^{N\times J} $ into latent code $\mathbf{z} \in \mathbb{R}^{K\times D}$ ($K \ll N$) in the latent space, while $\mathcal{D}$ is used to decode sample in latent space back to the real motion.
By using the Kullback-Leibler (KL) loss and the Mean Squared Error (MSE) loss for the training procedure, the motion VAE can provide a low-dimensional but representative latent space.

On the other hand, the diffusion model aims to generate a motion latent code in the VAE latent space according to the textual description, which is achieved by a reverse diffusion process that gradually transfers a random noise $\mathbf{n} \in \mathbb{R}^{K\times D}$ from Gaussian distribution to the motion latent code $\mathbf{z}_0 \in \mathbb{R}^{K\times D}$.
To train a denoiser $\epsilon_\theta$ for this reverse process, a forward diffusion process is required to successively add Gaussian noise onto $\mathbf{z}_0$ in a Markov chain manner, which can be formulated as: 
\begin{equation}
\begin{aligned}
q\left(\mathbf{z}_{1: T} \mid \mathbf{z}_0\right) & :=\prod_{t=1}^T q\left(\mathbf{z}_t \mid \mathbf{z}_{t-1}\right),
\end{aligned}
\end{equation}
\begin{equation}
\begin{aligned}
 q\left(\mathbf{z}_t \mid \mathbf{z}_{t-1}\right) & :=\mathcal{N}\left(\mathbf{z}_t ; \sqrt{1-\beta_t} \mathbf{z}_{t-1}, \beta_t \mathbf{I}\right),
\end{aligned}
\end{equation}
where $T$ is the total steps of the forward process and $\beta_t$ is a hyperparameter of the noising weight.
By using the reparameterization trick~\cite{kingma2014rt}, we can sample $\mathbf{z}_t$ from $q\left(\mathbf{z}_t \mid \mathbf{z}_{0}\right)$ at an arbitrary timestep $t$:
\begin{equation}
\label{equ:ztsample}
\mathbf{z}_t:=\sqrt{\bar{\alpha}_t} \mathbf{z}_0+\epsilon \sqrt{1-\bar{\alpha}_t}, 
 \;\;\;\; 
\epsilon \sim \mathcal{N}(\mathbf{0}, \mathbf{I}),
\end{equation}
where $\bar{\alpha}_t:=\prod_{s=1}^t \alpha_s$ and $\alpha_s:=1-\beta_s$. During training, given the noised data $\mathbf{z}_t$ and the textual description $\mathbf{w}$ as inputs, denoiser $\epsilon_\theta$ is expected to predict the noise $\epsilon$ added at $t$-th Markov step. The object function for the learning of $\epsilon_\theta$ only contains the MSE loss:
\begin{equation}
\label{equ:loss4diffusion}
\mathcal{L}:=\mathbb{E}_{\epsilon \sim \mathcal{N}(\mathbf{0}, \mathbf{I}), t \in[1, T]}\left[\| \epsilon-\epsilon_\theta\left(\mathbf{z}_t, \tau_\theta(\mathbf{w}), t \right) \|_2^2\right],
\end{equation}
where $\tau_\theta$ represents a pre-trained CLIP~\cite{nichol2021clip} text encoder which is used to extract the text embedding and is frozen during training. Furthermore, denoiser $\epsilon_\theta$ is trained by using classifier-free guidance~\cite{ho2021cfg}. Therefore, during inference, the predicted noise $\epsilon'$ is formulated as the linear combination of the conditional and unconditional predictions:
\begin{equation}
\epsilon':= \epsilon_\theta\left(\mathbf{z}_t, \varnothing, t \right) + g \cdot (\epsilon_\theta\left(\mathbf{z}_t, \tau_\theta(\mathbf{w}), t \right) - \epsilon_\theta\left(\mathbf{z}_t, \varnothing, t \right)),
\end{equation}
where $\varnothing$ represents a null-text input and $g$ is the hyperparameter of guidance scale.

\subsection{B2A-HDM}\label{sec3.2:b2ahdm}
Although using latent diffusion model can ease the learning of diffusion network, the low-dimensional latent space may be under representative (i.e., as shown in Fig~\ref{fig:dimension_comps}(a)) and thus constrains the generative upper bound of diffusion model, leading to detail-scarce motion synthesis. Directly increasing the dimension of the VAE latent space will make target distribution complex again and cause a significant performance drop in modality consistency (i.e., as illustrated in Fig~\ref{fig:dimension_comps}(c)). To boost the representation capacity of latent motion embedding without degrading the cross-modal mapping consistency, our B2A-HDM employs Basic Diffusion Model (BDM) to provide modality consistent generated results, which will be further processed by Advanced Diffusion Model (ADM) for detail-enhancing synthesis.

To be specific, our BDM and ADM are defined in the low-dimensional and high-dimensional latent space, respectively. 
To obtain BDM, a motion VAE $\mathcal{V}^l = \{\mathcal{E}^l,\mathcal{D}^l \}$ is trained on the raw motion distribution to obtain a low-dimensional latent space $\mathcal{W}^l$ (with latent code $\mathbf{z}^l \in \mathbb{R}^{K_1\times D}$). Then, a denoiser $\epsilon_\theta^l$ on $\mathcal{W}^l$ is trained to handle arbitrary $t$-th denoising step ($t \in [1, T]$), which will be used to conduct the reverse diffusion process from random noise $\mathbf{n}^l \in \mathbb{R}^{K_1\times D} $ during inference.
For ADM, motion VAE $\mathcal{V}^h = \{\mathcal{E}^h,\mathcal{D}^h \}$ is also required to obtain a  high-dimensional latent space $\mathcal{W}^h$ (with latent code $\mathbf{z}^h \in \mathbb{R}^{K_2\times D}$, $K_2 > K_1$). 
However, directly training a denoiser $\epsilon_\theta^h$ on $\mathcal{W}^h$ for arbitrary timestep $t$ is no-trivial and typically results in model degradation. 

To tackle this problem, B2A-HDM improves the common reverse diffusion process in two-folds. 
First, instead of using a single denoiser $\epsilon_\theta^h$ to conduct the whole reverse diffusion process, B2A-HDM applies BDM to provide the early $T^l$ steps denoising result for ADM, and ADM is only required to conduct the following $T-T^l$ denoising steps.
Since BDM is trained on $\mathcal{W}^l$ with lower distribution complexity and performs better in modality consistent synthesis, it can provide an intermediate denoising result with proper modality information. Therefore ADM is designed for the following detail-enhancing denoising process.
Furthermore, using a single denoiser $\epsilon_\theta^h$ for the remaining $T-T^l$ steps denoising process still remains challenging due to the high distribution complexity of $\mathcal{W}^h$ and the significant discrepancy of $\mathbf{z}_t^h$ in various timestep. To further ease the learning of denoiser on $\mathcal{W}^h$, our B2A-HDM assigns $k$ denoisers for ADM, in which each denoiser $\epsilon_\theta^{h_k}$ is only in charge of the specific interval of the denoising process in the high-dimensional latent space. 

During inference, the $T^l$ steps denoising result $\mathbf{z}_{T^l}^l$ from BDM can not used by ADM due to dimension inconsistency between $\mathbf{z}_{T^l}^l$ and $\mathbf{z}^h$ in $\mathcal{W}^h$. To address this problem, B2A-HDM first employs BDM to generate a motion sequence $\mathbf{s}^l$, which will be transferred into a latent code $\mathbf{\hat{z}}_0^h$ in $\mathcal{W}^h$. Then, B2A-HDM conducts $T-T^l$ steps forward diffusion to obtained intermediate denoising result for ADM.

We take $k=2$ as an example and show the complete reverse diffusion process as follows:

\begin{algorithm}
\caption{Reverse Diffusion Process of B2A-HDM}
\begin{algorithmic}[1]
\REQUIRE A textual description $\mathbf{w}$, a random seed $r$.   
\ENSURE A motion sequence $\mathbf{s}$.       
\STATE $\mathbf{z}_T^l \sim \mathcal{N}(\mathbf{0}, \mathbf{I})$, $\epsilon \sim \mathcal{N}(\mathbf{0}, \mathbf{I})$ unit Gaussian random variables with random seed $r$;        
\STATE $T^h \leftarrow T -T ^l$ denoising steps for ADM;

\FOR{$t = T, T-1, ..., 1$} 
\STATE $\mathbf{z}_{t-1}^l \leftarrow \epsilon_\theta^l(\mathbf{z}_t^l, \tau_\theta(\mathbf{w}), t)$; 
\ENDFOR
\STATE $\mathbf{s}^l \leftarrow \mathcal{D}^l(\mathbf{z}_0^l)$; \;\; $\mathbf{\hat{z}}_0^h \leftarrow \mathcal{E}^h(\mathbf{s}^l)$; \;\; $\mathbf{z}_{T^h}^h \leftarrow \sqrt{\bar{\alpha}_{T^h}} \mathbf{\hat{z}}_0^h+\epsilon \sqrt{1-\bar{\alpha}_{T^h}} $;
\FOR{$t = T^h, T^h-1, ..., 1$}
\IF {$t > \frac{T^h}{2}$}
\STATE $\mathbf{z}_{t-1}^h \leftarrow  \epsilon_\theta^{h_1}(\mathbf{z}_t^h, \tau_\theta(\mathbf{w}), t)$;
\ELSE
\STATE $\mathbf{z}_{t-1}^h \leftarrow  \epsilon_\theta^{h_2}(\mathbf{z}_t^h, \tau_\theta(\mathbf{w}), t)$;
\ENDIF
\ENDFOR
\STATE $\mathbf{s} \leftarrow \mathcal{D}^h(\mathbf{z}_0^h)$;
\RETURN $\mathbf{s}$
\end{algorithmic}
\end{algorithm}

\subsection{Training Details and Objective Functions}\label{sec3.3:train-test}
The training procedure for BDM and ADM are similar to that for the common latent diffusion model\cite{chen2023mld,rombach2021highresolution}, except for the training of the diffusion networks for ADM. 
Specifically, since denoisers in ADM are in charge of different denoising sub-processes, each denoiser is assigned a specific timestep interval during training.
Note that although each denoiser is independent of the others, we train all of them in a single training procedure, which enables us to conduct the complete reverse process during training and observe the performance change on the evaluation set.

During training VAE for BDM and ADM, the objective functions can be formulated as:
\begin{equation}
\mathcal{L}_{vae} = \lambda_{kl}\mathcal{L}_{kl} + \lambda_{mse}^{vae}\mathcal{L}_{mse}^{vae},
\end{equation}
where $\lambda_{kl}$ and $\lambda_{mse}^{vae}$ are the trade-off hyperparameters and are set to 1e-4 and 1.0, respectively.
When training the denoiser for BDM and ADM, only MSE loss $\mathcal{L}_{mse}^{\epsilon_\theta}$ in Equation~\ref{equ:loss4diffusion} is used. In addition, we introduce a timestep-aware loss weight $\lambda(t)$ for the MSE loss in BDM to increase the penalty for early denoising process learning. The timestep-aware MSE loss can be formulated as:
\begin{equation}
\mathcal{L}_{mse}^t = \lambda(t)\mathcal{L}_{mse}^{\epsilon}, \;\;\; \lambda(t) = (1-\bar{\alpha}_t) * w_1 + w_2 ,
\end{equation}
where $\bar{\alpha}_t$ is the diffusion parameter defined in Equation~\ref{equ:ztsample}, $w_1$ and $w_2$ are used to rescale the loss weight to a specific interval (i.e., $\lambda(t) \in [0.5,5] $) and are set to 4.5 and 0.5, respectively. In Sec.~\ref{sec:ab}, we will analyse the impact of using timestep-aware MSE loss for BDM, which will demonstrate improved performance in high-dimensional scenarios.

\begin{table*}[t]
    \def\arraystretch{1.2}
    \normalsize

    \tabcolsep 1.8pt
    \centering
    \begin{tabular}{c | l c c c c c c c}
        \toprule
        & \multirow{2}*{Method} & \multicolumn{3}{c}{R-Precision $\uparrow$}  & \multirow{2}*{FID $\downarrow$} & \multirow{2}*{MM-Dist $\downarrow$} & \multirow{2}*{Diversity $\rightarrow$} & \multirow{2}*{MModality $\uparrow$} \\
        \cmidrule{3-5}
        & & Top-1 & Top-2 & Top-3  \\
        \midrule

        \multirow{6}*{(a)}

        & \textbf{Real motion} & \et{0.511}{.003} & \et{0.703}{.003} & \et{0.797}{.002} & \et{0.002}{.000} & \et{2.974}{.008} & \et{9.503}{.065} & -  \\
        \cmidrule{2-9}
        & MDM \cite{tevet2023mdm} & \et{0.320}{.005} & \et{0.498}{.004} & \et{0.611}{.007} & \et{0.544}{.044} & \et{5.566}{.027} & \etbb{9.559}{.086} & \etr{2.799}{.072}   \\
        & MotionDiffuse \cite{zhang2022motiondiffuse} & \etbb{0.491}{.001} & \etbb{0.681}{.001} & \etbb{0.782}{.001} & \et{0.630}{.001} & \etbb{3.113}{.001} & \et{9.410}{.049} & \et{1.553}{.042}   
          \\
        & MLD \cite{chen2023mld} & \et{0.481}{.003} & \et{0.673}{.003} & \et{0.772}{.002} & \et{0.473}{.013} & \et{3.196}{.010} & \et{9.724}{.082} & \et{2.413}{.079}   
          \\
        & T2M-GPT \cite{zhang2023generating} & \etbb{0.491}{.003} & \et{0.680}{.003} & \et{0.775}{.002} & \etbb{0.116}{.004} & \et{3.118}{.011} & \et{9.761}{.081} & \et{1.856}{.011} 
          \\
        \cmidrule{2-9}
        & B2A-HDM (Ours) & \etr{0.511}{.002} & \etr{0.699}{.002} & \etr{0.791}{.002} & \etr{0.084}{.004} & \etr{3.020}{.010} & \etr{9.526}{.080} & \et{1.914}{.078} 
          \\
        \bottomrule
        
        \multirow{6}*{(b)}

        & \textbf{Real motion} & \et{0.424}{.005} & \et{0.649}{.006} & \et{0.779}{.006} & \et{0.031}{.004} & \et{2.788}{.012} & \et{11.08}{.097} & -  \\
        
        \cmidrule{2-9}
        & MDM \cite{tevet2023mdm} & \et{0.164}{.004} & \et{0.291}{.004} & \et{0.396}{.004} & \et{0.497}{.021} & \et{9.191}{.022} & \et{10.847}{.109} & \et{1.907}{.214}   \\
        & MotionDiffuse \cite{zhang2022motiondiffuse} & \etbb{0.417}{.004} & \et{0.621}{.004} & \et{0.739}{.004} & \et{1.954}{.062} & \etbb{2.958}{.005} & \etr{11.10}{.143} & \et{0.730}{.013}   
          \\
        & MLD \cite{chen2023mld} & \et{0.390}{.008} & \et{0.609}{.008} & \et{0.734}{.007} & \etbb{0.404}{.027} & \et{3.204}{.027} & \et{10.80}{.117} & \etbb{2.192}{.071}   
          \\
        & T2M-GPT \cite{zhang2023generating} & \et{0.416}{.006} & \etbb{0.627}{.006} & \etbb{0.745}{.006} & \et{0.514}{.029} & \et{3.007}{.023} & \etbb{10.921}{.108} & \et{1.570}{.039} 
          \\
        \cmidrule{2-9}
        & B2A-HDM (Ours) & \etr{0.436}{.006} & \etr{0.653}{.006} & \etr{0.773}{.005} & \etr{0.367}{.020} & \etr{2.946}{.024} & \et{10.86}{.124} & \et{1.291}{.047} 
          \\
        \bottomrule

    \end{tabular}
    \vspace{-2mm}
    \caption{Quantitative results on (a) HumanML3D~\cite{guo2022t2m} and (b) KIT-ML~\cite{Plappert2016}. \textcolor{red}{Red} and \textcolor{blue}{Blue} indicate the best and the second best result.}
    \vspace{-4mm}
    \label{tab:humanml_results}
\end{table*}

\begin{figure*}
  \centering
  \includegraphics[width=1.0\hsize]{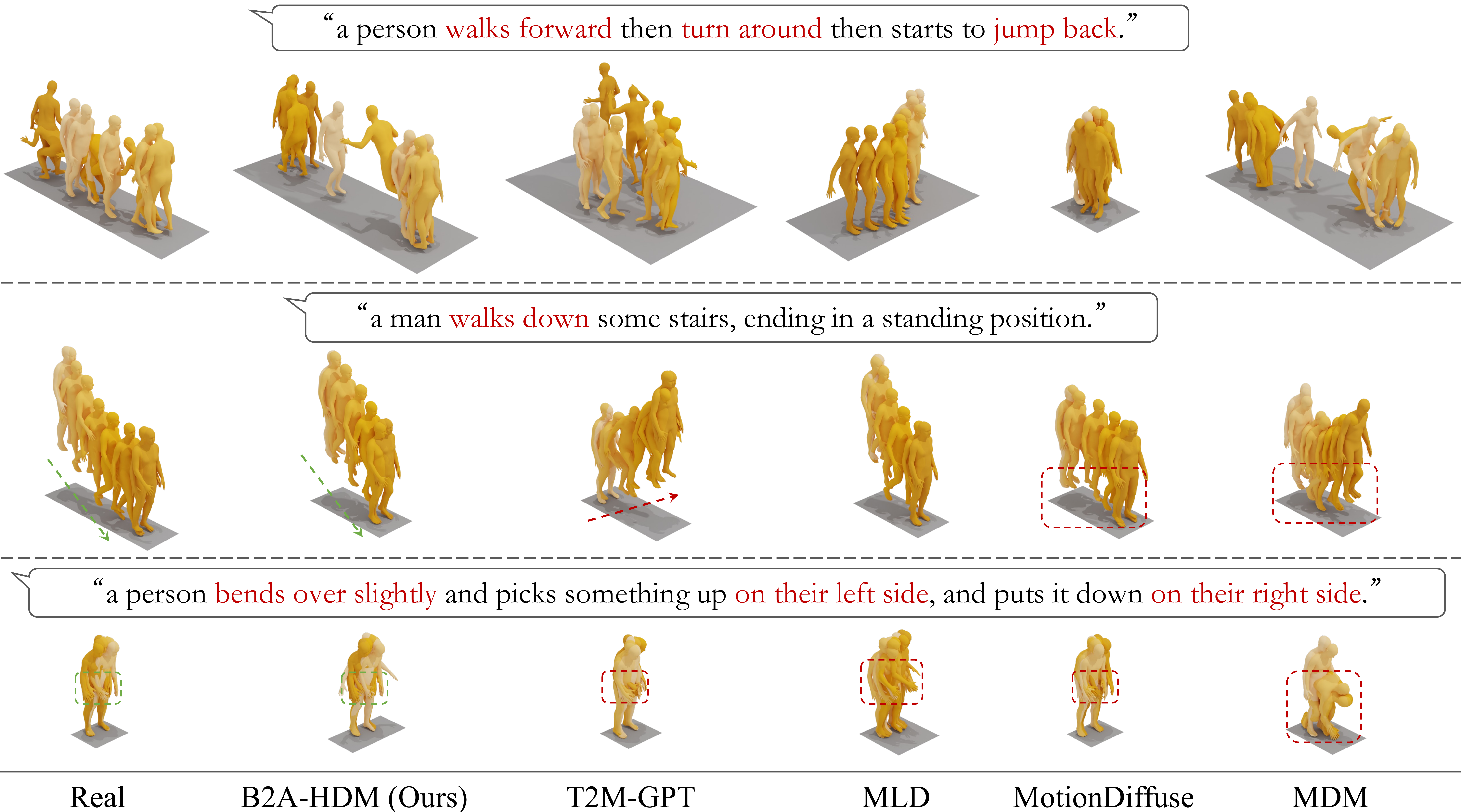}
  \vspace{-6mm}
  \caption{Qualitative comparisons on HumanML3D dataset~\cite{guo2022t2m}. The flow of time is represented by colors, with lighter shades indicating the past. Please zoom in for more details.}
   \vspace{-4mm}
  \label{fig:vis_comps}
\end{figure*}

\section{Experiments}
\noindent\textbf{Datasets.}
Our experiments are conducted on two publicly available benchmarks for text-to-motion synthesis, namely KIT-ML~\cite{Plappert2016} and HumanML3D~\cite{guo2022t2m}. Specifically, KIT-ML consists of 3,911 motion sequences with 12.5 FPS and 6,278 language annotations. 
HumanML3D contains 14,616 motion sequences with 20FPS and 44,970 textual descriptions. 
Regarding the data format, we follow~\cite{guo2022t2m} to use the redundant representation for each pose frame, which is composed of the local/global joint velocities, joint positions, joint rotations, and the foot contact binary labels.

\noindent\textbf{Baselines.}
We perform quantitative and qualitative comparisons with four most advanced text-to-motion synthesis methods, including MDM~\cite{tevet2023mdm}, MotionDiffuse~\cite{zhang2022motiondiffuse}, MLD~\cite{chen2023mld}, and T2M-GPT~\cite{zhang2023generating}. For these methods, we employ the official pre-trained models and strictly adhere to the official instructions to conduct text-to-motion synthesis.

\noindent\textbf{Evaluation Metrics.}
We employ five evaluation metrics originated from~\cite{guo2022t2m} to assess the performance of different methods. 
Specifically, FID~\cite{heusel2017fid} measures the distribution difference between the generated and real motion, which is widely used for realism evaluation. R-Precision and MM-Dist are designed to evaluate modality consistency, in which R-Precision calculates the Top-1/2/3 accuracy for the motion-to-text retrieval while MM-Dist calculates the Euclidean distances between the generated motion and its corresponding text. Diversity and MModality are designed for diversity evaluation, which are used to measure the variance of all generated motions and the variance of the particular generations for each text input, respectively.

\noindent\textbf{Implementation Details.}
The dimension of the latent space for BDM and ADM are $4 \times 256$ and $8 \times 256$, respectively. ADM is equipped with 2 denoisers. In spite of using different latent space, BDM and ADM share the same network architecture for the motion VAE encoder, decoder, and the diffusion denoiser. In line with MLD~\cite{chen2023mld}, all of the three modules are composed of 9 transformer layers with skip connection.
Our B2A-HDM is implemented using PyTorch~\cite{paszke2019pytorch} and both of motion VAE and diffusion denoiser are trained on 4 Tesla V100 GPUs.
During training, for both HumanML3D~\cite{guo2022t2m} and KIT-ML~\cite{Plappert2016} dataset, the batch size on each GPU is set to 96 and the all modules are trained by using AdamW~\cite{loshchilov2019adamw} optimizer with a fixed learning rate 1e-4. For HumanML3D, both VAE and denoiser are trained for 6,000 epochs, while for KIT-ML, the VAE and denoiser are trained for 25,000 epochs and 2,500 epochs, respectively.
\footnote{More details about the dataset, evaluation metrics, and network architecture are provided in the supplemental materials.}

\subsection{Quantitative Results}
The quantitative comparison of our B2A-HDM against the existing state-of-the-art methods on HumanML3D~\cite{guo2022t2m} and KIT-ML~\cite{Plappert2016} datasets are reported in Tab.~\ref{tab:humanml_results} (a) and (b), respectively. Both tables demonstrate that B2A-HDM outperforms other methods in terms of modality consistency and fidelity.
Specifically, B2A-HDM achieves the highest R-Precision (Top-1/2/3) and the lowest MM-Dist scores on both datasets, indicating that the generated results of B2A-HDM are more consistent with the input text than those of other methods. Moreover, B2A-HDM obtains the lowest FID score on both datasets, highlighting its superiority over other methods in realistic synthesis. 
It's worth noting that B2A-HDM is the only approach that consistently improves on the above three metrics, which further validates the effectiveness of the combination of basic and advanced diffusion models. 
Additionally, B2A-HDM achieves comparable Diversity and Modality scores on both datasets, demonstrating its ability for diverse generation.

\subsection{Qualitative Results}
Fig.~\ref{fig:vis_comps} shows a qualitative comparison of B2A-HDM against the existing SOTA methods on HumanML3D~\cite{guo2022t2m} dataset. The visual comparison illustrates B2A-HDM outperforms other methods in generating motion sequences with better modality consistency and detail preservation. For instance, in the first row of Fig.~\ref{fig:vis_comps}, MotionDiffuse~\cite{zhang2022motiondiffuse} and MDM~\cite{tevet2023mdm} fail to generate motion that coheres with the input text, while T2M-GPT~\cite{zhang2023generating} and MLD~\cite{chen2023mld} tend to overlook details of the hands.
In contrast, B2A-HDM generates motion sequences that conform to the input text and capture the fine-grained motion details in the hand region.
\footnote{More quantitative and qualitative results are provided in the supplemental materials.}

\begin{table}[t]
    \def\arraystretch{1.2}
    \normalsize

    \tabcolsep 3.0pt
    \centering
    \begin{tabular}{l| c c c c | c c}
        \toprule
        \multirow{2}*{Method}  & BD & AD  & LD-LS &  HD-LS &\multirow{2}{*}{\begin{tabular}[c]{@{}c@{}}R-P\\ Top-1\end{tabular} $\uparrow$} & \multirow{2}*{FID $\downarrow$} \\
        & No. & No. & Dim & Dim & \\
    \midrule
        BDM-4 & 1 & 0  & 4$\times$256  & -  &  0.505 & 0.284 \\
        ADM-8 & 0 & 1  & -  & 8$\times$256  &  0.481 & 0.171 \\
        B2A-HDM$\star$ & 3 & 0 &  4$\times$256  & -  &  0.508 & 0.220 \\
        B2A-HDM$\ast$ & 0 & 3 &  -  & 8$\times$256  &  0.490 & 0.225 \\
        \textbf{B2A-HDM} & 1 & 2  & 4$\times$256  & 8$\times$256  &  0.511 & 0.084 \\
        \bottomrule
    \end{tabular}
    \caption{Quantitative results of the ablation study with different configurations, in which BD/AD No., LD/HD-LS Dim refer to basic/advanced denoiser number and low/high-dimension latent space dimension, respectively}
    \label{tab:ab_abs}
  \vspace{-6mm}
\end{table}

\subsection{Ablation Study}\label{sec:ab}

\begin{figure}[t]
  \centering
  \includegraphics[width=1.0\hsize]{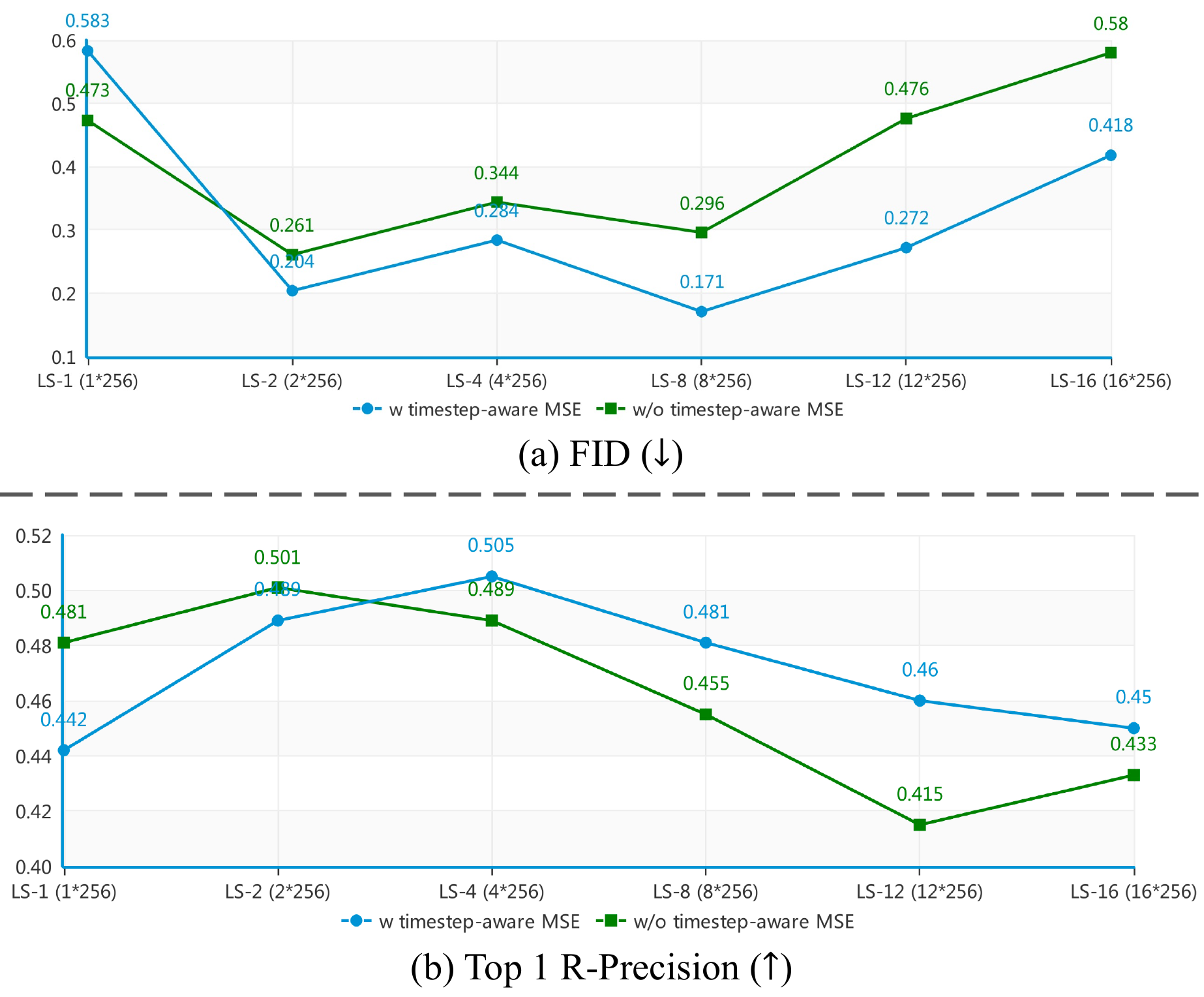}
  \vspace{-6mm}
  \caption{Impact of the timestep-aware MSE loss for BDMs in different latent space (LS).}
   \vspace{-4mm}
  \label{fig:vis_abs_lossweight}
\end{figure}

\noindent\textbf{Impact of the timestep-aware MSE loss.} 
As shown in Fig.~\ref{fig:vis_abs_lossweight}, when the dimension of the latent space is higher than $2 \times 256$, using the timestep-aware MSE loss consistently enhances the performance of diffusion model in terms of lower FID and higher Top-1 R-Precision, which highlights the ability of timestep-aware MSE loss to facilitate the learning of denoisers in high-dimensional latent spaces.

\begin{figure}[t]
  \centering
  \includegraphics[width=1.0\hsize]{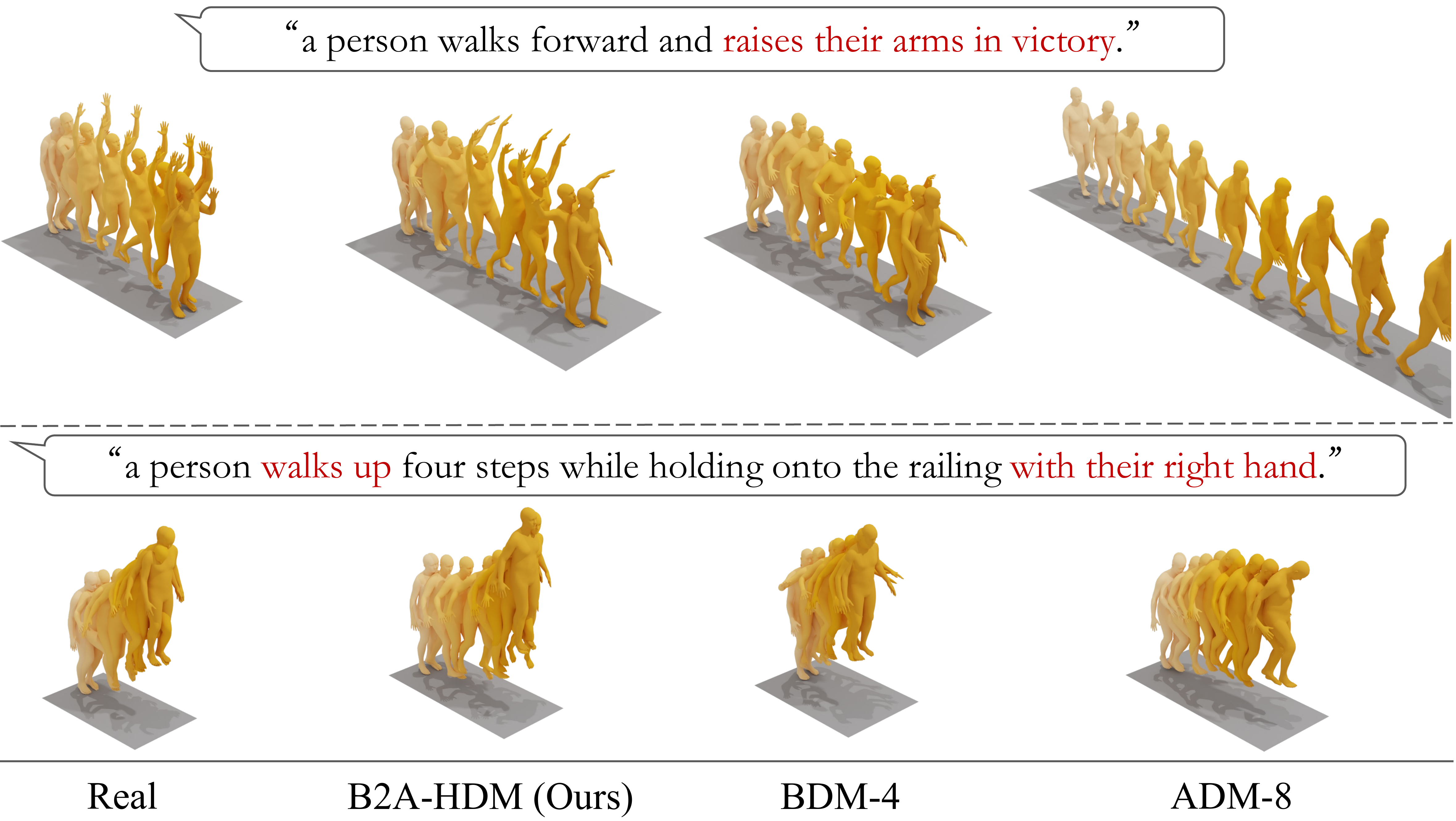}
  \vspace{-2mm}
  \caption{Ablation Study on the effectiveness of B2A-HDM. Please zoom in for more details.}
   \vspace{-2mm}
  \label{fig:vis_abs}
\end{figure}

\noindent\textbf{Effectiveness of B2A-HDM.} 
We compare B2A-HDM with BDM in a $4 \times 256$ latent space (BDM-4) and ADM in an $8 \times 256$ latent space (ADM-8). Additionally, we compare B2A-HDM with two variants (i.e., B2A-HDM$\star$ and B2A-HDM$\ast$) that separately include three denoisers in low-dimemsion and high-dimension latent space to demonstrate the effectiveness of combining basic and advanced diffusion models. 
As reported in Tab.~\ref{tab:ab_abs}, directly using BDM-4 or ADM-8 leads to higher FID or lower R-Precision. 
Although increasing the denoiser number enables B2A-HDM$\star$ and B2A-HDM$\ast$ to gain performance improvement against their single-denoiser counterparts (i.e., BDM-4 and ADM-8), they still fall short of B2A-HDM.
By collaborativly using basic and advanced diffusion models, B2A-HDM achieves best FID and R-Precision scores, which demonstrate the necessity and effectiveness of combining basic and advanced denoisers.
Moreover, Fig.~\ref{fig:vis_abs} shows that ADM-8 is prone to generate modality inconsistent motions while BDM-4 tends to ignore the motion details. In contrast, our B2A-HDM performs better in modality transformaion and detail preservation, which further validates the effectiveness of our method.


\section{Conclusion}
In this paper, we propose a novel Basic-to-Advanced Hierarchical Diffusion Model (B2A-HDM) for text-to-motion synthesis. Our B2A-HDM comprises a basic diffusion model (BDM) in low-dimensional latent space and a advanced diffusion model (ADM) with two denoisers in high-dimensional latent space, in which BDM are in charge of the modality-consistent denoising, whereas ADM is responsible for the following detail-enhancing denoising. In this way, our B2A-HDM can  fully leverage the generative potential of diffusion models to produce high-quality motion sequences that conform to the provided textual descriptions. Extensive experiments on two public text-to-motion benchmarks demonstrate the superiority of B2A-HDM over existing state-of-the-art methods, while ablation studies further validate the effectiveness of our approach.

\section{Acknowledgments}
This work was supported in part by National Key R\&D Program of China under Grant No.2020AAA0109700, Guangdong Outstanding Youth Fund (Grant No.2021B1515020061), National Natural Science Foundation of China (NSFC) under Grant No.61976233 and No.92270122, Mobility Grant Award under Grant No.M-0461, Shenzhen Science and Technology Program (Grant No.RCYX20200714114642083), Shenzhen Science and Technology Program (Grant No.GJHZ20220913142600001), Nansha Key RD Program under Grant No.2022ZD014 and Sun Yat-sen University under Grant No.22lgqb38 and 76160-12220011.

\bibliography{aaai24}

\clearpage


\section{Additional Experiment Details}\label{sec:experiment}

\subsection{Dataset Details}
Our experiments are conducted on two existing text-to-motion benchmarks, namely HumanML3D~\cite{guo2022t2m} and KIT-ML~\cite{Plappert2016}. 
HumanML3D dataset consists of 14,616 human motions and 44,970 text descriptions, in which each motion is annotated with 3 or 4 text descriptions and the average length of descriptions is approximately 12.
KIT-ML dataset consists of 3,911 human motions and 6,278 textual descriptions, in which each motion is annotated with 1 to 4 descriptions and the average length of descriptions is approximately 8.
We follow~\cite{guo2022t2m} to use the redundant representation for motion sequence in both datasets, which is also used by the recently state-of-the-art methods~\cite{rombach2021highresolution,tevet2023mdm,chen2023mld,zhang2023generating}.
To be specific, the representation for the $i$-th pose frame $p_i$ in a motion sequence $\mathbf{s}$ is composed of root angular velocity $\dot{r}^a \in \mathbb{R}$ along $Y$-axis, root linear velocities ($\dot{r}^x, \dot{r}^z \in \mathbb{R}$) on XZ-plane, root height $r^y \in \mathbb{R}$, local joints positions $\mathbf{j}^p \in \mathbb{R}^{3 j} $, velocities $\mathbf{j}^v \in \mathbb{R}^{3 j}$ and rorations $\mathbf{j}^r \in \mathbb{R}^{6 j}$ in the root space, and binary foot-ground concact features $\mathbf{c}^f \in \mathbb{R}^4$, in which $j$ represents the joint number and $\mathbf{c}^f$ is obtained by thresholding the heel and toe joint velocities. Therefore, the representation for $p_i$ can be denotated as $\{\dot{r}^a, \dot{r}^x, \dot{r}^z, r^y, \mathbf{j}^p, \mathbf{j}^v, \mathbf{j}^r, \mathbf{c}^f\}$.
Moreover, the skeleton of $p_i$ in HumamML3D and KIT-ML is with 22 and 21 joints respectively, while the represestntaion for $p_i$ in HumamML3D and KIT-ML is with dimension 263 and 251 respectively.

\subsection{Evaluation Metrics Details}
We provide more details about the evaluation metrics used in our experiments, which are derived from~\cite{guo2022t2m}. All of the metrics are calculated in the feature space, in which the motion feature and the text feature are extracted by the pre-trained motion encoder and text encoder in~\cite{guo2022t2m}, respectively. The motion encoder and text encoder are trained by employing a contrastive loss~\cite{hadsell2006cl} to enforce matched text-motion feature pairs to be as close as possible, and vice versa.

\noindent\textbf{Fréchet Inception Distance (FID).} FID~\cite{heusel2017fid} is widely used in the synthesis field to evaluate the resalism and fidelity of the synthesis, which is calculated between the distribution of the generated results and that of the real data. For our text-to-motion synthesis, FID can be formulated as follows:
\begin{equation}
\text { FID }=\left\|\mu_{gt}-\mu_{fake}\right\|-\operatorname{Tr}\left(\Sigma_{gt}+\Sigma_{fake}-2\sqrt{\Sigma_{gt} \Sigma_{fake}}\right),
\end{equation}
where $\mu_{gt}$/$\mu_{fake}$, $\Sigma_{gt}$/$\Sigma_{fake}$, and $\operatorname{Tr}$ represents the mean of the real/generated motion feature, covariance matrix for the real/generated motion feature, and the trace of a matrix, respectively.

\noindent\textbf{R-Precision.}
R-Precision measures the modality-consistency between the generated motions and provided texts. It calclates the Top-1/2/3 accuracy for motion-to-text retrieval. Specifically, for each generated motion, a candicated feature pool is constructed by using 1 feature of the matched text and 31 feature of the randomly selected mismatched texts.
Then, the retrival accuracy is calculated according to the Euclidean distance between the generated motion feature and the text feature from the feature pool.

\noindent\textbf{Multimodal Distance (MM-Dist).}
MM-Dist is another metric for modality-consistency evaluation. It calculates the Euclidean distance between the generated motion feature and its corresponding text feature.

\noindent\textbf{Diversity.}
Diversity measures the variance of the generated results.  Specifically, two sets with $N$ motions are randomly sampled from the generated results, respectively. Then, two motion feature sets (i.e., $\mathbf{F} = \{\mathbf{f}_1, ..., \mathbf{f}_N\}$ and $\mathbf{F}' = \{\mathbf{f}_1', ..., \mathbf{f}_N'\}$) are constructed from the above sampled sets, and Diversity is formulated as follows:
\begin{equation}
\text { Diversity }=\frac{1}{N} \sum_{i=1}^{N}\left\|\mathbf{f}_i-\mathbf{f}_i^{\prime}\right\|,
\end{equation}
where $N$ is set to 300 in our experiments.

\noindent\textbf{ Multimodality (MModality).}
MModality is another metric for diversity evaluation. However, it measures the diversity of the generated results for each textual description. Specifically, given a text set with size $C$, for the $c$-th description, two sets with $M$ generated motions are randomly sampled from all generated results for the $c$-th description. Then, two motion feature sets (i.e., $\mathbf{F}_c = \{\mathbf{f}_{c,1}, ..., \mathbf{f}_{c,M}\}$ and $\mathbf{F}_c' = \{\mathbf{f}_{c,1}', ..., \mathbf{f}_{c,M}'\}$) are constructed from the above sampled sets, and MModality is calculated as follows:
\begin{equation}
\text { MModality }=\frac{1}{C \times M} \sum_{c=1}^C \sum_{i=1}^{M}\left\|\mathbf{f}_{c, i}-\mathbf{f}_{c, i}^{\prime}\right\|,
\end{equation}
where $M$ is set to 10 in our experiments.


\section{Additional Experiment Results}\label{sec:result}

\subsection{Additional Quantitative Results}
In the main text, due to the space constrain, we only compare our B2A-HDM with four most advanced baselines. Here, we compare B2A-HDM with all typical baselines on HumanML3D~\cite{guo2022t2m} dataset and KIL-ML~\cite{Plappert2016} dataset, of which the quantitative comparison results are reported in Tab.~\ref{tab:humanml_results} and Tab.~\ref{tab:kit_results}, respectively.
Both tables shows that our B2A-HDM achieves best scores for R-Precision, MM-Dist, and FID, demonstrating that our B2A-HDM outperforms other methods in terms of modality consistenct and fidelity. Besides, B2A-HDM obtain comparable Diversity and Modality scores on both datasets, demonstrating its ability for diverse generation.

Furthermore,  We conduct human evaluation to compare our B2A-HDM with the other state-of-the-art methods according to their synthesis quality.
Specifically, we invite 19 volunteers to complete the questionaire that consists of 15 assignments. 
For each assignment, given a textual description, the volunteers are asked to select the optimal human motion (i.e., most consistent with the text and realistic) out of five options, which are generated by our B2A-HDM, MDM~\cite{tevet2023mdm}, MotionDiffuse~\cite{zhang2022motiondiffuse}, MLD~\cite{chen2023mld}, and T2M-GPT~\cite{zhang2023generating}, respectively. 
Besides, the order of the generated motion is each assignment is randomly shuffled.
The human evaluation results are shown in Tab.~\ref{tab:human_evaluation}, which demonstrate the superiority of B2A-HDM over the state-of-the-art methods.

\begin{table}
    \small
    \tabcolsep 1.5pt
    \centering
    \begin{tabular}{l | c c c c c}
    \toprule
     Method & B2A-HDM & MDM & MotionDiffuse & MLD & T2M-GPT \\
    \midrule
    HE Score & 47.0\% & 3.50\% & 11.6\% & 25.6\% & 12.3\% \\ 
    \bottomrule
    \end{tabular}
    \caption{Human Evaluation (HE) Results on HumanML3D dataset~\cite{guo2022t2m}}
    \label{tab:human_evaluation}
    \vspace{-4mm}
\end{table}

\subsection{Additional Visual Results}
We provide additional generated results and visual comparisons among our proposed B2A-HDM and the other state-of-the-art methods(i.e., MDM~\cite{tevet2023mdm}, MotionDiffuse~\cite{zhang2022motiondiffuse}, MLD~\cite{chen2023mld} and T2M-GPT~\cite{zhang2023generating}) in the project repository: https://github.com/xiezhy6/B2A-HDM. Please refer to the video for more details.

\subsection{Resource Consumption Comparison}
In this section, we compare the resource consumption (i.e., parameter consumption and computation consumption) among our B2A-HDM and four most advanced baselines (i.e., MDM, MotionDiffuse, MLD, and T2M-GPT), of which the comparison results are shown in Tab.~\ref{tab:consumption}.

For parameter consumption, since the text encoder used in different methods varies, we separately compare the parameter consumption of the motion generator and the full model. 
For computation consumption, we compare the consumption of different methods in terms of the GPU memory consumption and test time consumption when generating a single motion under the same experiment setting (i.e., using same hardware/software environment, receiving the same text input). 

As shown in Tab.~\ref{tab:consumption}(a), although B2A-HDM contains multi-denoisers, the parameter consumption for its generator is only 60.9M, which is about a quarter of that for the existing SOTA method T2M-GPT.
As shown in Tab.~\ref{tab:consumption}(b), for GPU memory consumption, except for MDM, B2A-HDM and the other three baseline occupy similar GPU memory.
For test time consumption, the average time cost for B2A-HDM is 0.658s, which is closed to the least time cost 0.274s (from MLD) and significantly superior over that of MotionDiffuse and MDM.

\begin{table*}[t]
    \small
    \tabcolsep 8.0pt
    \centering
    \begin{tabular}{c | l | c c c c c}
    \toprule
        & Method & B2A-HDM & MLD & MotionDiffuse & MDM & T2M-GPT \\
        \midrule
        \multirow{2}*{(a)} 
        & Parameter Consumption (Generator only)	 & 60.9M &	26.3M &	87.1M &	17.8M	& 247.4M \\
        & Parameter Consumption (Full model) &	487.9M &	453.3M &	238.1M &	168.8M &	398.4M \\
        \midrule
        \multirow{2}*{(b)} 
         & GPU Memory Consumption & 	3GB & 	2.9GB &	2.8GB	& 1.4GB &	2.5GB \\
         & Test Time Consumption &	0.658s &	0.274s &	15.899s & 	29.661s &	0.430s \\
    \bottomrule
    \end{tabular}
    \caption{Resource consumption comparisons.}
    \label{tab:consumption}
\end{table*}

\subsection{Model Hyperparameter Analysis}
In this section, we compare B2A-HDM with its variants to comprehensively analyze the influence of model hyperparameters, which include the number of advanced denoiser in ADM (named Advanced Denoiser Number in Tab.~\ref{tab:hyperparameters}), the rate of the reverse diffusion process that ADM is responsible for (named Advanced Diffusion Rate in Tab.~\ref{tab:hyperparameters}), and the dimension of the low-dimensional latent space (named LD-LS in Tab.~\ref{tab:hyperparameters}) and high-dimensional latent space (named HD-LS in Tab.~\ref{tab:hyperparameters}). 
Furthermore, for each hyperparameter, we provide an empirical guidance on value selection for training B2A-HDM on new dataset.

According to Tab.~\ref{tab:hyperparameters}(b), when the advanced diffusion rate changes from 25\% to 75\%, both of the R-Precision score and FID socre of the variants fluctuate within a small interval, in which the R-Precision scores are similar to that of full B2A-HDM while the FID scores are inferior to that of full B2A-HDM. Therefore, to obtain lower FID score and guarantee the realism of the generated result, we suggest to use higher advanced diffusion rate (e.g., larger than 80\%) for training B2A-HDM on new dataset.

According to Tab.~\ref{tab:hyperparameters}(c), B2A-HDM$\dagger$-1 (variant with one advanced denoiser) obtains the worst FID score, while B2A-HDM$\dagger$-2 and B2A-HDM$\dagger$-3 (variants with three or four advanced denoisers) obtain significant improvement for FID score, which illustrates that more advanced denoisers can ease the learning of denoiser in high dimension latent space. However, since using two advanced denosiers already obtains the optimal performance, we suggest to use such setting for training B2A-HDM on new dataset.

According to Tab.~\ref{tab:hyperparameters}(d), when fixing the dimension of low-dimension latent space, decreasing the dimension of the high-dimension latent space obtains performance improvement. On the other hand, when fixing the dimension of high-dimension latent space, increasing the dimension of low-dimension latent space obtains performance improvement. Actually, to determine the latent space for the basic and advanced denosier, we first test single-denoiser latent diffusion model in different latent space, and choose the latent space where the denoiser obtains best performance as the low-dimension latent space. Then, we choose another latent space with higher dimension as the high-dimension latent space.

It is worth noting that, we directly employ the optimal hyperparameters for HumanML3D dataset to train B2A-HDM on KIT-ML dataset, obtaining a model consistently outperforms baseline methods in terms of R-Precsion, MM-Dist and FID. Therefore, we believe that the default hyperparameters for HumanML3D dataset can provide a favorable initial setting for training B2A-HDM on other datasets.

\begin{table*}[t]
    \small

    \tabcolsep 2pt
    \centering
    \begin{tabular}{l c c c c c c c}
        \toprule
        \multirow{2}*{Method} & \multicolumn{3}{c}{R-Precision $\uparrow$}  & \multirow{2}*{FID $\downarrow$} & \multirow{2}*{MM-Dist $\downarrow$} & \multirow{2}*{Diversity $\rightarrow$} & \multirow{2}*{MModality $\uparrow$} \\
        \cmidrule{2-4}
        & Top-1 & Top-2 & Top-3  \\
        \midrule

        \textbf{Real motion} & \et{0.511}{.003} & \et{0.703}{.003} & \et{0.797}{.002} & \et{0.002}{.000} & \et{2.974}{.008} & \et{9.503}{.065} & -  \\
    \midrule
        Seq2Seq \cite{lin2018seq2seq} & \et{0.180}{.002} & \et{0.300}{.002} & \et{0.396}{.002} & \et{11.75}{.035} & \et{5.529}{.007} & \et{6.223}{.061}  & -  \\

        Language2Pose \cite{ahuja2019l2p} & \et{0.246}{.002} & \et{0.387}{.002} & \et{0.486}{.002} & \et{11.02}{.046} & \et{5.296}{.008} & \et{7.676}{.058} & -  \\

        Text2Gesture \cite{bhattacharya2021t2g} & \et{0.165}{.001} & \et{0.267}{.002} & \et{0.345}{.002} & \et{5.012}{.030} & \et{6.030}{.008} & \et{6.409}{.071} & -  \\

        Hier \cite{ghosh2021hier} & \et{0.301}{.002} & \et{0.425}{.002} & \et{0.552}{.004} & \et{6.532}{.024} & \et{5.012}{.018} & \et{8.332}{.042} & -  \\

        MoCoGAN \cite{tulyakov2018mocogan} & \et{0.037}{.000} & \et{0.072}{.001} & \et{0.106}{.001} & \et{94.41}{.021} & \et{9.643}{.006} & \et{0.462}{.008} & \et{0.019}{.000}  \\

        Dance2Music \cite{lee2019d2m} & \et{0.033}{.000} & \et{0.065}{.001} & \et{0.097}{.001} & \et{66.98}{.016} & \et{8.116}{.006} & \et{0.725}{.011} & \et{0.043}{.001}  \\

        TM2T \cite{guo2022tm2t} & \et{0.424}{.003} & \et{0.618}{.003} & \et{0.729}{.002} & \et{1.501}{.017} & \et{3.467}{.011} & \et{8.589}{.076} & \etbb{2.424}{.093}  \\

        T2M \cite{guo2022t2m} & \et{0.457}{.002} & \et{0.639}{.003} & \et{0.740}{.003} & \et{1.067}{.002} & \et{3.340}{.008} & \et{9.188}{.002} & \et{2.090}{.083}  \\
        MDM \cite{tevet2023mdm} & \et{0.320}{.005} & \et{0.498}{.004} & \et{0.611}{.007} & \et{0.544}{.044} & \et{5.566}{.027} & \etbb{9.559}{.086} & \etr{2.799}{.072}   \\
        MotionDiffuse \cite{zhang2022motiondiffuse} & \etbb{0.491}{.001} & \etbb{0.681}{.001} & \etbb{0.782}{.001} & \et{0.630}{.001} & \etbb{3.113}{.001} & \et{9.410}{.049} & \et{1.553}{.042}   \\
        MLD \cite{chen2023mld} & \et{0.481}{.003} & \et{0.673}{.003} & \et{0.772}{.002} & \et{0.473}{.013} & \et{3.196}{.010} & \et{9.724}{.082} & \et{2.413}{.079}   \\
        T2M-GPT \cite{zhang2023generating} & \etbb{0.491}{.003} & \et{0.680}{.003} & \et{0.775}{.002} & \etbb{0.116}{.004} & \et{3.118}{.011} & \et{9.761}{.081} & \et{1.856}{.011} \\
    \midrule
        B2A-HDM (Ours) & \etr{0.511}{.002} & \etr{0.699}{.002} & \etr{0.791}{.002} & \etr{0.084}{.004} & \etr{3.020}{.010} & \etr{9.526}{.080} & \et{1.914}{.078} 
          \\
        \bottomrule
    \end{tabular}
    \caption{Quantitative comparisons on HumanML3D dataset~\cite{guo2022t2m}. \textcolor{red}{Red} and \textcolor{blue}{Blue} indicate the best and the second best result.}
    \label{tab:humanml_results}
\end{table*}

\begin{table*}[t]    
    \small

    \tabcolsep 2pt
    \centering
    \begin{tabular}{l c c c c c c c}
        \toprule
        \multirow{2}*{Method} & \multicolumn{3}{c}{R-Precision $\uparrow$}  & \multirow{2}*{FID $\downarrow$} & \multirow{2}*{MM-Dist $\downarrow$} & \multirow{2}*{Diversity $\rightarrow$} & \multirow{2}*{MModality $\uparrow$} \\
        \cmidrule{2-4}
        & Top-1 & Top-2 & Top-3  \\
        \midrule

        \textbf{Real motion} & \et{0.424}{.005} & \et{0.649}{.006} & \et{0.779}{.006} & \et{0.031}{.004} & \et{2.788}{.012} & \et{11.08}{.097} & -  \\
    \midrule
        Seq2Seq \cite{lin2018seq2seq} & \et{0.103}{.003} & \et{0.178}{.005} & \et{0.241}{.006} & \et{24.86}{.348} & \et{7.960}{.031} & \et{6.744}{.106}  & -  \\

        Language2Pose \cite{ahuja2019l2p} & \et{0.221}{.005} & \et{0.373}{.004} & \et{0.483}{.005} & \et{6.545}{.072} & \et{5.147}{.030} & \et{9.073}{.100} & -  \\

        Text2Gesture \cite{bhattacharya2021t2g} & \et{0.156}{.004} & \et{0.255}{.004} & \et{0.338}{.005} & \et{12.12}{.183} & \et{6.964}{.029} & \et{9.334}{.079} & -  \\

        Hier \cite{ghosh2021hier} & \et{0.255}{.006} & \et{0.432}{.007} & \et{0.531}{.007} & \et{5.203}{.107} & \et{4.986}{.027} & \et{9.563}{.072} & -  \\

        MoCoGAN \cite{tulyakov2018mocogan} & \et{0.022}{.002} & \et{0.042}{.003} & \et{0.063}{.003} & \et{82.69}{.242} & \et{10.47}{.012} & \et{3.091}{.043} & \et{0.250}{.009}  \\

        Dance2Music \cite{lee2019d2m} & \et{0.031}{.002} & \et{0.058}{.002} & \et{0.086}{.003} & \et{115.4}{.240} & \et{10.40}{.016} & \et{0.241}{.004} & \et{0.062}{.002}  \\

        TM2T \cite{guo2022tm2t} & \et{0.280}{.005} & \et{0.463}{.006} & \et{0.587}{.005} & \et{3.599}{.153} & \et{4.591}{.026} & \et{9.473}{.117} & \etr{3.292}{.081}  \\

        T2M \cite{guo2022t2m} & \et{0.361}{.006} & \et{0.559}{.007} & \et{0.681}{.007} & \et{3.022}{.107} & \et{3.488}{.028} & \et{10.72}{.145} & \et{2.052}{.107}  \\
        MDM \cite{tevet2023mdm} & \et{0.164}{.004} & \et{0.291}{.004} & \et{0.396}{.004} & \et{0.497}{.021} & \et{9.191}{.022} & \et{10.847}{.109} & \et{1.907}{.214}   \\
        MotionDiffuse \cite{zhang2022motiondiffuse} & \etbb{0.417}{.004} & \et{0.621}{.004} & \et{0.739}{.004} & \et{1.954}{.062} & \etbb{2.958}{.005} & \etr{11.10}{.143} & \et{0.730}{.013}  \\
        MLD \cite{chen2023mld} & \et{0.390}{.008} & \et{0.609}{.008} & \et{0.734}{.007} & \etbb{0.404}{.027} & \et{3.204}{.027} & \et{10.80}{.117} & \etbb{2.192}{.071}  \\
        T2M-GPT \cite{zhang2023generating} & \et{0.416}{.006} & \etbb{0.627}{.006} & \etbb{0.745}{.006} & \et{0.514}{.029} & \et{3.007}{.023} & \etbb{10.921}{.108} & \et{1.570}{.039} \\
    \midrule
        B2A-HDM (Ours) & \etr{0.436}{.006} & \etr{0.653}{.006} & \etr{0.773}{.005} & \etr{0.367}{.020} & \etr{2.946}{.024} & \et{10.86}{.124} & \et{1.291}{.047} 
          \\
        \bottomrule
    \end{tabular}
    \caption{Quantitative comparisons on KIT-ML dataset~\cite{Plappert2016}. \textcolor{red}{Red} and \textcolor{blue}{Blue} indicate the best and the second best result.}
    \label{tab:kit_results}
\end{table*}

\begin{table*}[t]
    \small

    \tabcolsep 8.0pt
    \centering
    \begin{tabular}{c|c| c c c c | c c}
        \toprule
        & \multirow{2}*{Method}  & Advanced & Advanced & LD-LS &  HD-LS &\multirow{2}{*}{\begin{tabular}[c]{@{}c@{}}R-Precision\\ Top-1\end{tabular} $\uparrow$} & \multirow{2}*{FID $\downarrow$} \\
        & & Denoiser Number & Diffusion Rate & Dimension & Dimension & \\
    \midrule
        (a) & \textbf{B2A-HDM}  & 2 & 95\% & 4$\times$256  & 8$\times$256  &  \et{0.511}{.002} & \etb{0.084}{.004} \\
    \midrule
        \multirow{3}*{(b)}
        & B2A-HDM$\diamond$-1  & 2 & 25\% & 4$\times$256  & 8$\times$256  &  \et{0.509}{.003} & \et{0.242}{.006} \\
        & B2A-HDM$\diamond$-2  & 2 & 50\% & 4$\times$256  & 8$\times$256  &  \et{0.507}{.002} & \et{0.277}{.008} \\
        & B2A-HDM$\diamond$-3  & 2 & 75\% & 4$\times$256  & 8$\times$256  &  \etb{0.514}{.002} & \et{0.288}{.008} \\
    \midrule
        \multirow{3}*{(c)}
        & B2A-HDM$\dagger$-1  & 1 & 95\% & 4$\times$256  & 8$\times$256  &  \et{0.505}{.002} & \et{0.315}{.009} \\
        & B2A-HDM$\dagger$-2  & 3 & 95\% & 4$\times$256  & 8$\times$256  &  \et{0.505}{.003} & \et{0.130}{.006} \\
        & B2A-HDM$\dagger$-3  & 4 & 95\% & 4$\times$256  & 8$\times$256  &  \et{0.511}{.003} & \et{0.139}{.007} \\
    \midrule
        \multirow{4}*{(d)}
        & B2A-HDM$\ddagger$-1  & 2 & 95\% & 4$\times$256  & 12$\times$256  &  \etb{0.514}{.003} & \et{0.123}{.005} \\
        & B2A-HDM$\ddagger$-2  & 2 & 95\% & 4$\times$256  & 16$\times$256  &  \et{0.506}{.003} & \et{0.125}{.005} \\
        & B2A-HDM$\ddagger$-3  & 2 & 95\% & 1$\times$256  & 8$\times$256  &  \et{0.508}{.003} & \et{0.153}{.006} \\
        & B2A-HDM$\ddagger$-4  & 2 & 95\% & 2$\times$256  & 8$\times$256  &  \et{0.511}{.003} & \et{0.103}{.005} \\

        \bottomrule
    \end{tabular}
    \caption{Quantitative results of different B2A-HDM variants with various hyperparameter configurations.}
    \label{tab:hyperparameters}
\end{table*}

\begin{table*}[t]
    \def\arraystretch{1.05}
    \small

    \tabcolsep 2pt
    \centering
    \begin{tabular}{l l}
        \toprule
        Modules & Architecture \\
        \midrule
        PE Layer & (query\_pos\_encoder): PositionEmbeddingLearned1D() \\
         & (query\_pos\_decoder): PositionEmbeddingLearned1D() \\
        \midrule
        Mapping & (skel\_embedding): Linear(in\_feat=263, out\_feat=256, bias=True) \\
        Layer & (final\_layer): Linear(in\_feat=256, out\_feat=263, bias=True) \\
        \midrule
        VAE $\mathcal{E}$ & (encoder): SkipTransformerEncoder( \\
        & \ \ \ \ (norm): LayerNorm((256,), eps=1e-05, elementwise\_affine=True) \\
        & \ \ \ \ (input\_blocks): ModuleList( \\
        & \ \ \ \ \ \ \ \ (0): 4 $\times$ TransformerEncoderLayer( \\
        & \ \ \ \ \ \ \ \ \ \ \ \ \ (self\_attn): MultiheadAttention(NonDynQuanLinear(in\_feat=256, out\_feat=256, bias=True)) \\
        & \ \ \ \ \ \ \ \ \ \ \ \ \ (linear1): Linear(in\_feat=256, out\_feat=1024, bias=True) \\
        & \ \ \ \ \ \ \ \ \ \ \ \ \ (dropout): Dropout(p=0.1, inplace=False) \\
        & \ \ \ \ \ \ \ \ \ \ \ \ \ (linear2): Linear(in\_feat=1024, out\_feat=256, bias=True) \\
        & \ \ \ \ \ \ \ \ \ \ \ \ \ (norms): 2 $\times$ LayerNorm((256,), eps=1e-05, elementwise\_affine=True) \\
        & \ \ \ \ \ \ \ \ \ \ \ \ \ (dropouts): 2 $\times$ Dropout(p=0.1, inplace=False) \\
        & \ \ \ \ (middle\_block): TransformerEncoderLayer( \\
        & \ \ \ \ \ \ \ \ (self\_attn): MultiheadAttention(NonDynQuanLinear(in\_feat=256, out\_feat=256, bias=True)) \\
        & \ \ \ \ \ \ \ \ (linear1): Linear(in\_feat=256, out\_feat=1024, bias=True) \\
        & \ \ \ \ \ \ \ \ (dropout): Dropout(p=0.1, inplace=False) \\
        & \ \ \ \ \ \ \ \ (linear2): Linear(in\_feat=1024, out\_feat=256, bias=True) \\
        & \ \ \ \ \ \ \ \ (norms): 2 $\times$ LayerNorm((256,), eps=1e-05, elementwise\_affine=True) \\
        & \ \ \ \ \ \ \ \ (dropouts): 2 $\times$ Dropout(p=0.1, inplace=False) \\
        & \ \ \ \ (output\_blocks): ModuleList( \\
        & \ \ \ \ \ \ \ \ (0): 4 $\times$ TransformerEncoderLayer( \\
        & \ \ \ \ \ \ \ \ \ \ \ \ (self\_attn): MultiheadAttention(NonDynQuanLinear(in\_feat=256, out\_feat=256, bias=True)) \\
        & \ \ \ \ \ \ \ \ \ \ \ \ (linear1): Linear(in\_feat=256, out\_feat=1024, bias=True) \\
        & \ \ \ \ \ \ \ \ \ \ \ \ (dropout): Dropout(p=0.1, inplace=False) \\
        & \ \ \ \ \ \ \ \ \ \ \ \ (linear2): Linear(in\_feat=1024, out\_feat=256, bias=True) \\
        & \ \ \ \ \ \ \ \ \ \ \ \ (norms): 2 $\times$ LayerNorm((256,), eps=1e-05, elementwise\_affine=True) \\
        & \ \ \ \ \ \ \ \ \ \ \ \ (dropout): 2 $\times$ Dropout(p=0.1, inplace=False) \\
        & \ \ \ \ (linear\_blocks): ModuleList( (0): 4 $\times$ Linear(in\_feat=512, out\_feat=256, bias=True)) \\
        \midrule
        VAE $\mathcal{D}$ & (decoder): SkipTransformerDecoder( \\
        & \ \ \ \ (norm): LayerNorm((256,), eps=1e-05, elementwise\_affine=True) \\
        & \ \ \ \ (input\_blocks): ModuleList( \\
        & \ \ \ \ \ \ \ \ (0): 4 $\times$ TransformerDecoderLayer( \\
        & \ \ \ \ \ \ \ \ \ \ \ \ \ (attns): 2 $\times$ MultiheadAttention(NonDynQuanLinear(in\_feat=256, out\_feat=256, bias=True)) \\
        & \ \ \ \ \ \ \ \ \ \ \ \ \ (linear1): Linear(in\_feat=256, out\_feat=1024, bias=True) \\
        & \ \ \ \ \ \ \ \ \ \ \ \ \ (dropout): Dropout(p=0.1, inplace=False) \\
        & \ \ \ \ \ \ \ \ \ \ \ \ \ (linear2): Linear(in\_feat=1024, out\_feat=256, bias=True) \\
        & \ \ \ \ \ \ \ \ \ \ \ \ \ (norms): 3 $\times$ LayerNorm((256,), eps=1e-05, elementwise\_affine=True) \\
        & \ \ \ \ \ \ \ \ \ \ \ \ \ (dropouts): 3 $\times$ Dropout(p=0.1, inplace=False) \\
        & \ \ \ \ (middle\_block): TransformerDecoderLayer( \\
        & \ \ \ \ \ \ \ \ (attns): 2 $\times$ MultiheadAttention(NonDynQuanLinear(in\_feat=256, out\_feat=256, bias=True)) \\
        & \ \ \ \ \ \ \ \ (linear1): Linear(in\_feat=256, out\_feat=1024, bias=True) \\
        & \ \ \ \ \ \ \ \ (dropout): Dropout(p=0.1, inplace=False) \\
        & \ \ \ \ \ \ \ \ (linear2): Linear(in\_feat=1024, out\_feat=256, bias=True) \\
        & \ \ \ \ \ \ \ \ (norms): 3 $\times$ LayerNorm((256,), eps=1e-05, elementwise\_affine=True) \\
        & \ \ \ \ \ \ \ \ (dropouts): 3 $\times$ Dropout(p=0.1, inplace=False) \\
        & \ \ \ \ (output\_blocks): ModuleList( \\
        & \ \ \ \ \ \ \ \ (0): 4 $\times$ TransformerDecoderLayer( \\
        & \ \ \ \ \ \ \ \ \ \ \ \ (attns): 2 $\times$ MultiheadAttention(NonDynQuanLinear(in\_feat=256, out\_feat=256, bias=True)) \\
        & \ \ \ \ \ \ \ \ \ \ \ \ (linear1): Linear(in\_feat=256, out\_feat=1024, bias=True) \\
        & \ \ \ \ \ \ \ \ \ \ \ \ (dropout): Dropout(p=0.1, inplace=False) \\
        & \ \ \ \ \ \ \ \ \ \ \ \ (linear2): Linear(in\_feat=1024, out\_feat=256, bias=True) \\
        & \ \ \ \ \ \ \ \ \ \ \ \ (norms): 3 $\times$ LayerNorm((256,), eps=1e-05, elementwise\_affine=True) \\
        & \ \ \ \ \ \ \ \ \ \ \ \ (dropout): 3 $\times$ Dropout(p=0.1, inplace=False) \\
        & \ \ \ \ (linear\_blocks): ModuleList( (0): 4 $\times$ Linear(in\_feat=512, out\_feat=256, bias=True)) \\
        \bottomrule
    \end{tabular}
    \caption{Architecture of motion VAE $\mathcal{V} = \{\mathcal{E}, \mathcal{D}\}$.}
    \label{tab:vae}
\end{table*}

\begin{table*}[t]
    \def\arraystretch{1.2}
    \small

    \tabcolsep 1pt
    \centering
    \begin{tabular}{l l}
        \toprule
        Modules & Architecture \\
        \midrule
        Timestep & (time\_proj): Timesteps() \\
        \midrule
        PE Layer & (query\_pos): PositionEmbeddingLearned1D() \\
        & (mem\_pos): PositionEmbeddingLearned1D() \\
        \midrule
        Mapping & (time\_embedding): TimestepEmbedding( \\
        Layer & \ \ \ \ (linear\_1): Linear(in\_feat=768, out\_feat=256, bias=True) \\
        & \ \ \ \ (act): SiLU() \\
        & \ \ \ \ (linear\_2): Linear(in\_feat=256, out\_feat=256, bias=True)) \\
        & (emb\_proj): Sequential( \\
        & \ \ \ \ (0): ReLU() \\
        & \ \ \ \ (1): Linear(in\_feat=768, out\_feat=256, bias=True)) \\
        \midrule
        denoiser $\epsilon_{\theta}$ & (encoder): SkipTransformerEncoder( \\
        & \ \ \ \ (norm): LayerNorm((256,), eps=1e-05, elementwise\_affine=True) \\
        & \ \ \ \ (input\_blocks): ModuleList( \\
        & \ \ \ \ \ \ \ \ (0): 4 $\times$ TransformerEncoderLayer( \\
        & \ \ \ \ \ \ \ \ \ \ \ \ \ (self\_attn): MultiheadAttention(NonDynQuanLinear(in\_feat=256, out\_feat=256, bias=True)) \\
        & \ \ \ \ \ \ \ \ \ \ \ \ \ (linear1): Linear(in\_feat=256, out\_feat=1024, bias=True) \\
        & \ \ \ \ \ \ \ \ \ \ \ \ \ (dropout): Dropout(p=0.1, inplace=False) \\
        & \ \ \ \ \ \ \ \ \ \ \ \ \ (linear2): Linear(in\_feat=1024, out\_feat=256, bias=True) \\
        & \ \ \ \ \ \ \ \ \ \ \ \ \ (norms): 2 $\times$ LayerNorm((256,), eps=1e-05, elementwise\_affine=True) \\
        & \ \ \ \ \ \ \ \ \ \ \ \ \ (dropouts): 2 $\times$ Dropout(p=0.1, inplace=False) \\
        & \ \ \ \ (middle\_block): TransformerEncoderLayer( \\
        & \ \ \ \ \ \ \ \ (self\_attn): MultiheadAttention(NonDynQuanLinear(in\_feat=256, out\_feat=256, bias=True)) \\
        & \ \ \ \ \ \ \ \ (linear1): Linear(in\_feat=256, out\_feat=1024, bias=True) \\
        & \ \ \ \ \ \ \ \ (dropout): Dropout(p=0.1, inplace=False) \\
        & \ \ \ \ \ \ \ \ (linear2): Linear(in\_feat=1024, out\_feat=256, bias=True) \\
        & \ \ \ \ \ \ \ \ (norms): 2 $\times$ LayerNorm((256,), eps=1e-05, elementwise\_affine=True) \\
        & \ \ \ \ \ \ \ \ (dropouts): 2 $\times$ Dropout(p=0.1, inplace=False) \\
        & \ \ \ \ (output\_blocks): ModuleList( \\
        & \ \ \ \ \ \ \ \ (0): 4 $\times$ TransformerEncoderLayer( \\
        & \ \ \ \ \ \ \ \ \ \ \ \ (self\_attn): MultiheadAttention(NonDynQuanLinear(in\_feat=256, out\_feat=256, bias=True)) \\
        & \ \ \ \ \ \ \ \ \ \ \ \ (linear1): Linear(in\_feat=256, out\_feat=1024, bias=True) \\
        & \ \ \ \ \ \ \ \ \ \ \ \ (dropout): Dropout(p=0.1, inplace=False) \\
        & \ \ \ \ \ \ \ \ \ \ \ \ (linear2): Linear(in\_feat=1024, out\_feat=256, bias=True) \\
        & \ \ \ \ \ \ \ \ \ \ \ \ (norms): 2 $\times$ LayerNorm((256,), eps=1e-05, elementwise\_affine=True) \\
        & \ \ \ \ \ \ \ \ \ \ \ \ (dropout): 2 $\times$ Dropout(p=0.1, inplace=False) \\
        & \ \ \ \ (linear\_blocks): ModuleList( (0): 4 $\times$ Linear(in\_feat=512, out\_feat=256, bias=True)) \\
        \bottomrule
    \end{tabular}
    \caption{Architecture of diffusion network $\epsilon_{\theta}$.}
    \label{tab:denoiser}
\end{table*}

\section{Architecture Details of B2A-HDM}\label{sec:network}
Our proposed B2A-HDM consists of Basic Diffusion Model (BDM) in low-dimensional latent space and Advanced Diffusion Model (ADM) in high-dimensional latent space.
In spite of using different latent spaces, VAE $\mathcal{V} = \{\mathcal{E}, \mathcal{D}\}$ and denoiser $\epsilon_{\theta}$ of BDM and ADM share the same network architecture. We illustrate the architecture of $\mathcal{V}$ and $\epsilon_{\theta}$ in Tab.~\ref{tab:vae} and Tab.~\ref{tab:denoiser}, respectively.

\section{Broader Impacts and Limitations}\label{sec:limitation}
\noindent\textbf{Broader Impacts.}
Like most generative models, our B2A-HDM might be applied to malicious manipulations. Specifically, it may be collaboratively used with the motion transfer algorithm to transfer weird motions onto specific person without permission. However, some advanced methods like forensics analysis and other manipulation detection methods can largely alleviate this negative impact.

\noindent\textbf{Limitations.}
Since our B2A-HDM is trained on the publicly available datasets~\cite{guo2022t2m,lee2019d2m} with limited amout of text-annotated motions, it owns its limitation to handle textual descriptions with arbitrary style. For example, if the provided descriptions are with extremently concrete or brief style, our proposed B2A-HDM may generate inferior results. To improve the model generalization for various text styles, we could resort to Large Language Model like (ChatGPT-3.5/ChatGPT-4) to enrich the style of textual descriptions in the motion dataset and employ the enriched descriptions for model training.
On the other hand, like most of the existing methods, our B2A-HDM primarily focuses on the generation of articulated human body and neglects the details in the face and hands. We believe that synthesizing motion with accurate facial expressions and hand movements would add valuable meaning to our work.

\end{document}